\documentclass{article}

\usepackage{PRIMEarxiv}

\usepackage[utf8]{inputenc} 
\usepackage[T1]{fontenc}    
\usepackage{hyperref}       
\usepackage{url}            
\usepackage{booktabs}       
\usepackage{amsfonts}       
\usepackage{nicefrac}       
\usepackage{microtype}      
\usepackage{lipsum}
\usepackage{longtable} 
\usepackage{array}
\usepackage{amsmath}
\usepackage{graphicx}
\usepackage{enumitem}
\usepackage{tikz}
\usetikzlibrary{arrows.meta} 
\usepackage{fancyhdr}       
\usepackage{graphicx}       
\graphicspath{{media/}}     

\title{ A Multi-Faceted Evaluation Framework for Assessing Synthetic Data Generated by Large Language Models
}

\author{
  Yefeng Yuan, Yuhong Liu \\
  Santa Clara University \\
  Santa Clara, CA\\
  \texttt{\{yyuan4, yhliu\}@scu.edu} \\
   \And
  Liang Cheng \\
  eBay Inc. \\
  San Jose, CA \\
  \texttt{liacheng@ebay.com} \\
}

\begin{document}
\maketitle

\begin{abstract}
The rapid advancements in generative AI and large language models (LLMs) have opened up new avenues for producing synthetic data, particularly in the realm of structured tabular formats, such as product reviews. Despite the potential benefits, concerns regarding privacy leakage have surfaced, especially when personal information is utilized in the training datasets. In addition, there is an absence of a comprehensive evaluation framework capable of quantitatively measuring the quality of the generated synthetic data and their utility for downstream tasks. In response to this gap, we introduce SynEval \footnote{https://github.com/SCU-TrustworthyAI/SynEval}, an open-source evaluation framework designed to assess the fidelity, utility, and privacy preservation of synthetically generated tabular data via a suite of diverse evaluation metrics. We validate the efficacy of our proposed framework - SynEval - by applying it to synthetic product review data generated by three state-of-the-art LLMs: ChatGPT, Claude, and Llama. Our experimental findings illuminate the trade-offs between various evaluation metrics in the context of synthetic data generation. Furthermore, SynEval stands as a critical instrument for researchers and practitioners engaged with synthetic tabular data, empowering them to judiciously determine the suitability of the generated data for their specific applications, with an emphasis on upholding user privacy.

\end{abstract}

\keywords{Synthetic data generation \and Privacy \and Evaluation Framework \and Large Language Model \and Responsive AI}

\section{Introduction}
The proliferation of high-quality data serves as the cornerstone propelling the current advancements in artificial intelligence (AI). Nonetheless, real-world datasets frequently suffer from noise and  imbalances, impeding the performance of AI models trained on such data. For example, in the context of product reviews, there is often a lack of data representing marginalized communities, such as disabled users. This data scarcity can lead to biased and inaccurate models that fail to capture the diverse needs and experiences of all users. Also, collecting and labeling real data can be expensive, time consuming, and error-prone. For instance, it is reported that, on average, data labeling can cost organizations \$2.3 million per year and 16 weeks to perform supervised learning on a new project\cite{noteSynAI}. Furthermore, privacy regulations make it even more difficult for companies to utilize their customers’ data even if such data is available at their premises. For example, the fines imposed by GDPR in 2023 alone have already exceeded the total amount of fines from 2019 to 2021 \cite{porter2023}, with a record-breaking \$1.28 billion imposed for Meta’s failure in complying with GDPR during its data transfer from EU to the U.S. for Facebook services \cite{notePriManager}. 

As a result, rather than directly utilizing real data, generating synthetic data that can sufficiently represent the statistical properties of real data while preserving individual user privacy is gaining extensive attention. Compared to real data, synthetic data can be generated in large quantities much faster than the same amount of real data can be collected, and can be more easily manipulated to allow more precise and controlled training and testing of AI models \cite{Savage2023,Dilmegani2023}. According to Gartner, it is estimated that synthetic data will play a dominant role in AI models by 2030, and 89\% of tech executives believed that synthetic data is a key for companies to stay competitive \cite{VentureBeat2021}. 

Due to its numerous benefits, synthetic data generation has attracted interests from a wide range of companies, such as Google, Microsoft, Amazon, Facebook, Nvidia \cite{Metinko2022,Yoon2019, Yoon2022}. For example, Google Cloud recently announced partnership with Gretel for generating anonymized and safe synthetic data \cite{Golshan2023}. Microsoft has collaborated with Harvard to develop synthetic data generation tools for responsive AI \cite{Barron2021}. Amazon has developed in-house synthetic data tools for training computer vision models (AWS, 2023). Facebook acquired a synthetic data generator startup AI.Reverie \cite{Wiggers2021}. Nvidia is creating a synthetic data generation engine for training AI networks \cite{Nvidia2023}. Recent advancements in generative AI, such as Generative Adversarial Networks (GANs) \cite{goodfellow2014generative} and Large Language Models (LLMs) \cite{vaswani2017attention}, have shown promising results in generating realistic synthetic data.

However, existing evaluation frameworks for synthetic data often focus on one or two specific aspects. There is a lack of a comprehensive evaluation framework that merges multiple perspectives and offers a holistic suite of metrics for the assessment of synthetic data. This poses a significant challenge in the adoption and trustworthiness of synthetic data generation techniques. Therefore, in this work, we propose SynEval, a comprehensive framework to rigorously evaluate the effectiveness of synthetic data generation from these critical dimensions: 
\begin{itemize}
    \item \textbf{Fidelity}: This aspect focuses on the degree to which synthetic data replicates the statistical characteristics of the original dataset. 
    \item \textbf{Utility}: The utility of synthetic data is determined by its effectiveness in facilitating various downstream machine learning tasks. This involves evaluating whether models trained on synthetic data perform comparably to those trained on real data when tested on validation or real-world datasets. 
    \item \textbf{Privacy}: Privacy protection is paramount to ensure that the synthetic data does not reveal sensitive information about individuals in the original dataset. Privacy can be assessed through techniques like re-identification risk analysis and other privacy-preserving metrics. 
\end{itemize}

The development of robust evaluation metrics and frameworks like SynEval is critical for the advancement of synthetic data generation methods. Without such tools, it is challenging to gauge the quality and safety of synthetic data, which can hinder its adoption in sensitive domains such as ecommerce, healthcare, and finance.

The major contributions of this work are as follows:
\begin{itemize}[topsep=0pt,partopsep=0pt,itemsep=0pt,parsep=0pt]
\item We propose a multi-faceted evaluation framework that integrates data fidelity, utility, and privacy evaluation with a comprehensive set of evaluation metrics to provide a holistic assessment of synthetically generated data.
\item We demonstrate the effectiveness of the proposed framework by applying it to synthetic product review data generated by three prominent LLMs: ChatGPT, Claude, and Llama.
\item We provide insights and recommendations based on the evaluation results, highlighting the strengths and limitations of each large language model in generating high-quality, useful, and privacy-preserving synthetic tabular data.
\end{itemize}

By addressing the research gap and providing a comprehensive evaluation framework, the proposed work contributes to the advancement of synthetic data generation techniques and promotes the responsible and trustworthy use of synthetic data in various applications. The proposed framework serves as a valuable tool for researchers and practitioners to assess the effectiveness of synthetic tabular data generated by LLMs and make informed decisions regarding their deployment in real-world scenarios.

\section{Related Work}
\label{sec:relatedwork}

\subsection{Synthetic Data Generation}
Synthetic data generation has gained significant attention in recent years to address privacy concerns, data scarcity, and data imbalance issues. Various approaches have been proposed for generating synthetic data, including statistical models \cite{soltana2017synthetic}, generative adversarial networks (GANs) \cite{jordon2018pate}, variational autoencoders (VAEs) \cite{wan2017variational}, and more recently, LLMs \cite{li2023synthetic}. 
Statistical models, such as Bayesian networks \cite{heckerman2008tutorial} and copulas \cite{frees1998understanding}, have been used to generate synthetic data by learning the underlying probability distributions of the real data. These models often require domain knowledge and explicit modeling of the data dependencies, which can be challenging for complex datasets.  
GANs have emerged as a popular approach for generating realistic synthetic data by training a generator network to produce samples that are indistinguishable from real data. GANs have been successfully applied to various domains, including image generation \cite{bao2017cvae}, text generation \cite{zhang2017adversarial}, and tabular data generation \cite{xu2019modeling}. However, training GANs can be challenging, and ensuring the stability and convergence of the training process remains an active area of research \cite{qi2024efficient}.
VAEs, on the other hand, learn a latent representation of the data and generate synthetic samples by sampling from the learned latent space. VAEs have been used to generate synthetic tabular data and have shown promising results in preserving the statistical properties of the real data. However, the generated samples may lack diversity and realism \cite{wan2017variational}.

Recently, LLMs, such as GPT-3 \cite{floridi2020gpt}, Claude \cite{wu2023comparative}, and Llama \cite{touvron2023llama} have demonstrated remarkable capabilities in generating coherent and realistic text. These models have been pre-trained on vast amounts of diverse text data and can be fine-tuned for specific tasks, including synthetic data generation \cite{tang2023does}.

\subsection{Quantitative Evaluation Metrics}

With diverse approaches for generating synthetic data, it is essential to quantitatively evaluate the quality of generated synthetic data, which can help compare different generation approaches and ensure diverse, balanced, and privacy-preserving data for downstream tasks. To this end, various evaluation metrics have been proposed. 

Statistical similarity metrics, such as Jensen-Shannon divergence \cite{nguyen2015non}, and maximum mean discrepancy \cite{dziugaite2015training}, have been introduced to quantify the statistical differences between synthetic data and real data. These measures capture the overall statistical properties of the data but may not account for the specific characteristics of individual features or feature correlations. To evaluate the utility of synthetic data for downstream machine learning tasks, typical machine learning performance metrics, such as accuracy, precision, recall, and F1-score, are often used by existing literature \cite{snoke2018general}. By training machine learning models on synthetic data and evaluating their performance on real data, researchers can assess the extent to which the synthetic data retains the inference capability of the real data. However, these metrics may not capture the nuances and biases presented in the synthetic data. Privacy evaluation frameworks, such as differential privacy(DP) \cite{dwork2009complexity}, k-anonymity \cite{sweeney2002k}, and l-diversity \cite{machanavajjhala2007diversity}, have been proposed to assess the privacy risks associated with synthetic data. However, these techniques often assume a specific level of attacker knowledge and are inadequate to evaluate general privacy risks. Differential privacy has emerged as a strong mathematical framework that provides quantitative privacy guarantees by adding carefully calibrated noise to the data or query results.
Nevertheless, evaluating whether the synthetically generated data satisfies DP requirements is challenging due to the high computational overhead. Therefore, in the proposed SynEval framework, we evaluate the privacy of synthetic data by checking its robustness against privacy attacks, specifically the Membership Inference Attacks (MIA) \cite{shokri2017membership}.


Existing evaluation frameworks often focus on specific aspects of synthetic data quality and may not provide a comprehensive assessment of the generated data. Moreover, evaluating synthetic tabular data generated by LLMs presents unique challenges due to the complex nature of the data, which often comprises a combination of discrete categorical variables and free-form text fields, along with intricate relationships and dependencies between these elements that must be preserved to ensure the utility and integrity of the data, while also maintaining semantic consistency, coherence, and the protection of sensitive information to mitigate privacy risks. Therefore, a comprehensive evaluation framework that integrates multiple perspectives and is tailored to the characteristics of synthetic tabular data generated by LLMs is needed.

\section{Proposed Scheme}


\begin{figure}[t] 
  \centering 
  \includegraphics[width=0.8\textwidth]{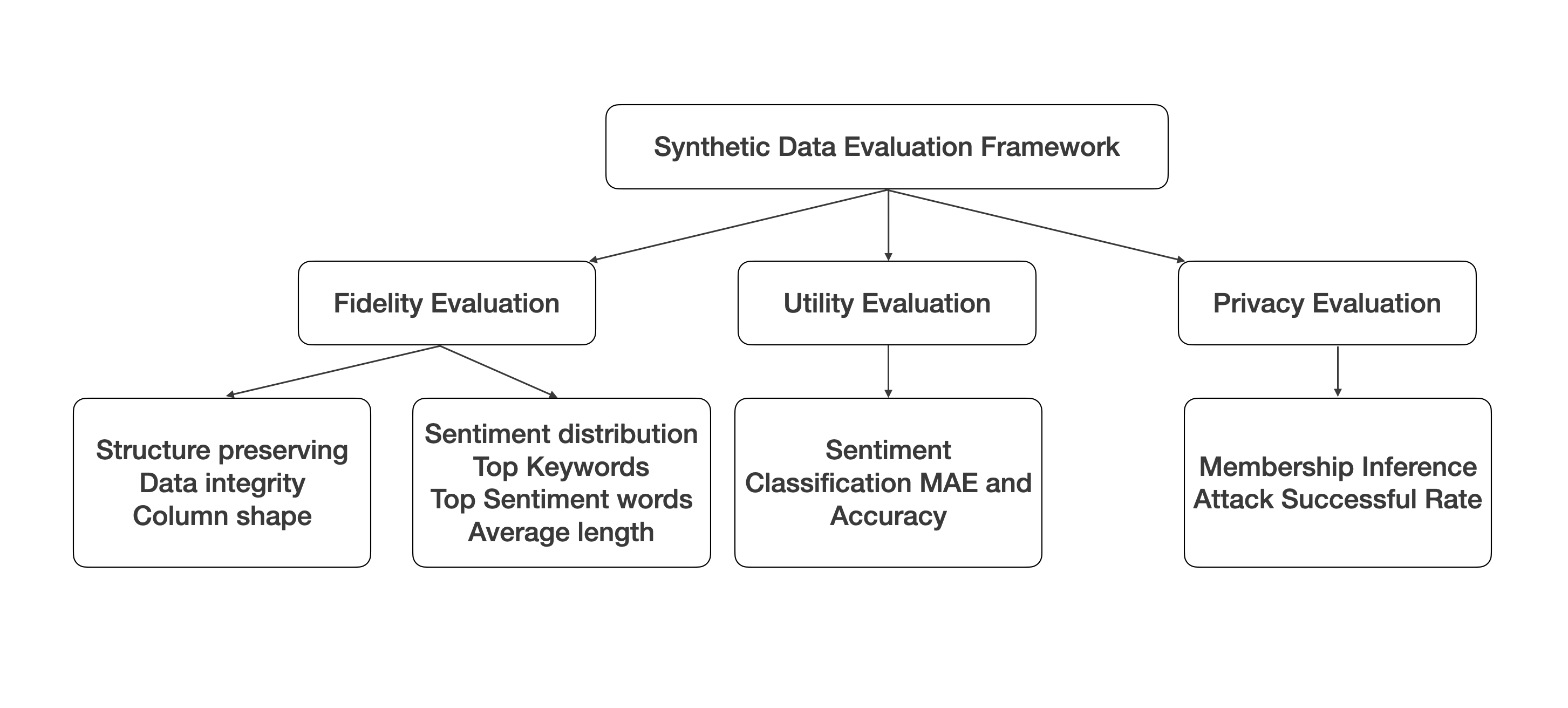} 
    \vspace{-9mm}
  \caption{Framework Overview} 
  \label{fig:example} 
\vspace{-6mm}
\end{figure}

Figure 1 presents an overview of the proposed framework, which comprises three evaluations: fidelity, utility, and privacy. The fidelity evaluation employs metrics such as structure preservation, data integrity, and column shape calculations for non-text tabular data. Additionally, it includes metrics for review text, such as sentiment distribution, top keywords and sentiment-related words, and average length. The utility evaluation is determined by calculating the accuracy of downstream sentiment classification. Lastly, the privacy evaluation is based on the success rate of membership inference attacks.

\subsection{Fidelity Evaluation}
Data fidelity evaluation is a crucial component of the proposed framework, as it assesses the degree to which the synthetic tabular data resembles the real data. Ensuring high fidelity is essential for building trust in the generated synthetic data and enabling its effective use in various applications, such as model training, testing, and decision-making. 

In particular, the proposed framework emphasizes the importance of comprehensive data fidelity evaluation, considering multiple aspects of data similarity and quality. Different from existing literature, the proposed data fidelity evaluation includes not only simple statistical comparisons but also an in-depth analysis of the relationships, dependencies, and contextual information present in the data. 


First, for non-text tabular data, we introduce the \textbf{Structure Preserving Score} (SPS) to evaluate whether the synthetic data retains the same column names and order as the real data. This assessment involves identifying all column names in the real (r) and synthetic (s) datasets and calculating the overlap between these columns to derive a fidelity score.

\begin{equation}
    \text{SPS} = \frac{\left| r \cap s \right|}{\left| r \cup s \right|} \cite{patki2016synthetic}
\end{equation}

Next, we introduce \textbf{Integrity Score} (IS) to assess the integrity of the non-text tabular data. Specifically, the computation of IS depends on whether the data is continuous or discrete. The overall IS is the average IS of continuous and discrete data. For continuous data, such as timestamps, we assess whether synthetic columns adhere to the minimum and maximum values defined by the corresponding real columns. For discrete data, such as ratings and verified purchases, the proposed approach checks if synthetic columns maintain the same category values as those in the real data. It is crucial that the synthetic data does not introduce new category values absent from the real dataset. The metric first recognizes set of unique categories from a real column (Cr), then it calculates the percentage of synthetic data (s) that are found in the set compared to all synthetic data points in this column.
\begin{equation}
    \text{IS} = \frac{\left| s, s \in C_r \right|}{\left| s \right|} \cite{patki2016synthetic}
\end{equation}


The third evaluation criterion for non-text tabular data assesses whether the synthetic data accurately capture the distribution shape of each column. \textbf{Column shape} refers to its overall distribution pattern. To evaluate this, we employ different methods depending on the data type. For continuous numerical values, such as helpful vote, we utilize the Kolmogorov-Smirnov \cite{berger2014kolmogorov} statistic to measure the similarity in marginal distributions between the synthetic and real columns. For discrete data, such as verified purchases, we apply the Total Variation Distance (TVD) \cite{patki2016synthetic} to compare the synthetic and real columns. The overall column shapes score is then calculated as the average across all columns.

To rigorously evaluate the fidelity of textual review data, we have developed a text analysis tool. This tool begins by assessing the \textbf{sentiment distribution} of each review, categorizing them as positive, neutral, or negative. We then compare the distribution of these sentiment categories between the real and synthetic data to identify and analyze dominant sentiments, ensuring that the synthetic data accurately reflects the emotional tone of the real reviews.

Furthermore, we conduct a detailed lexical analysis by extracting the \textbf{top three keywords} and \textbf{sentiment-related words} from the real dataset and synthetic data. This step involves not only identifying these pivotal words but also quantifying their frequency to establish a benchmark for comparison. We then evaluate how well these keywords and sentiment indicators are replicated in the synthetic data, which provides insight into the synthetic model’s ability to capture and reproduce the key thematic elements of the real text.

Additionally, we measure the \textbf{average length} of the reviews in both the real and synthetic data. This metric is crucial as it helps determine whether the synthetic data can maintain the same level of detail as the real reviews, which is important for applications where text length may influence the perceived quality or informativeness of the content.

Through these multifaceted analytical approaches, we ensure a comprehensive evaluation of the synthetic textual data, gauging its fidelity not only in mimicking the overt sentiments expressed in the reviews but also in preserving the underlying thematic and stylistic nuances that characterize the real dataset.

\subsection{Utility Evaluation}
The utility evaluation component of the proposed framework plays a crucial role in assessing the usefulness of the generated synthetic tabular data for downstream machine learning tasks. It is essential to ensure that the synthetic data retains the predictive power and informativeness of the real data, enabling the development of accurate and reliable machine learning models. To evaluate the utility of the synthetic tabular data, we propose adopting a framework named TSTR (Train-Synthetic-Test-Real), which offers a robust measure of synthetic data quality by assessing how well machine learning models trained on synthetic data perform on previously unseen real data.

In the context of product reviews, a common machine learning task is sentiment classification, which involves predicting the sentiment (positive, negative, or neutral) of a given review based on its text content. We begin by extracting (review, rating) pairs from both the original and synthetic datasets to train sentiment classification models using logistic regression. The models are then validated on a separate set of untouched real data to ensure robustness and prevent overfitting. \textbf{Accuracy} is determined by how closely the predicted ratings align with the actual ratings from the review texts, incorporating a tolerance threshold to account for minor variations. Additionally, we calculate the \textbf{Mean Absolute Error (MAE)} across all test instances to quantify the average prediction error, providing a clear metric of the models’ performance in real-world scenarios.

By comparing models trained on synthetic data to those trained on real data, we assess whether synthetic data preserves the utility of its real counterparts. This utility evaluation offers an in-depth analysis of the synthetic tabular data's effectiveness, extending beyond basic statistical comparisons. It directly gauges the performance of synthetic data in real-world machine learning applications, providing valuable insights into its quality and utility.

\subsection{Privacy Evaluation}

In the privacy evaluation of the proposed work, we employ Membership Inference Attacks (MIA) \cite{carlini2022membership} to assess whether synthetic datasets can reliably mimic the privacy attributes of real datasets, hence protecting the anonymity of the underlying data sources. This approach not only tests the robustness of the privacy-preserving measures but also quantifies the risk of sensitive information leakage through synthetic data.

We initiate the process by loading and preprocessing both real and synthetic datasets, which include data labeled as from 'members' (real dataset) and 'non-members' (synthetic dataset). For each dataset, categorical features such as product IDs, parent product IDs, and user IDs are encoded using label encoding. The datasets processed include real member data, real non-member data, and synthetic data generated by different language models. For the features' transformation, we implement a column transformer that processes different types of data.

Using the prepared datasets, we construct a combined dataset of real members and a randomly selected half of all synthetic datasets. This combined dataset serves as our training set. The test set comprises real non-member data combined with the remaining half of the synthetic data. 

A Random Forest Classifier is trained on this mixed dataset within a pipeline that includes the preliminary feature transformation. We assess the classifier’s performance by calculating \textbf{successful rate} of predictions. Higher successful rate in distinguishing members from non-members suggest greater potential for privacy leakage, indicating that the synthetic data may be too closely replicating identifiable patterns present in the real data.

This comprehensive scheme for privacy evaluation using MIA models thus plays a crucial role in our assessment of the security and applicability of synthetic datasets generated by state-of-the-art language models, guiding us in making informed decisions about their deployment in sensitive or privacy-conscious environments.

\section{Experiments}
\subsection{Synthetic Data Generation Scenario}
In this work, our primary focus is on the generation of synthetic tabular data with text reviews. Tabular data encompasses a structured format that is pivotal in numerous analytical scenarios across various domains. The ability to accurately and efficiently generate synthetic tabular data not only aids in enhancing data privacy but also facilitates the testing and development of new analytical models where real data may be sparse or sensitive.

We specifically concentrate on online review data for several reasons. First, such data is abundant and rich in diverse types of information, making it an ideal candidate for robust synthetic data generation experiments. Online reviews, as seen on platforms like Amazon, not only influence consumer behavior but also provide a wealth of information that can be mined for insights into user satisfaction and product quality.

We use the Amazon product review dataset \cite{hou2024bridging}, which contains millions of product reviews from various categories, such as electronics, books, and clothing, as the source dataset. User review data includes various data fields, which are structured below. 


\begin{itemize}[label={\textbullet}]
\item \textbf{Rating}: Numerical rating of the product, ranging from 1.0 to 5.0, indicating the overall customer satisfaction.
\item \textbf{Title}: Brief title of the user review, summarizing the user's opinion.
\item \textbf{Text}: Detailed text body of the user review, providing in-depth feedback and commentary on the product.
\item \textbf{Images}: Visual content uploaded by users post-purchase, showcasing the product from a consumer's perspective.
\item \textbf{ASIN}: The Amazon Standard Identification Number, a unique code assigned to each product listed on Amazon.
\item \textbf{Parent ASIN}: The identifier for the parent product, under which variations (such as size, color, etc.) are listed.
\item \textbf{User ID}: Identifier for the user who posted the review, ensuring traceability and authenticity.
\item \textbf{Timestamp}: The exact time when the review was posted, expressed in Unix time format.
\item \textbf{Helpful Vote}: Count of the number of times other users found the review helpful.
\item \textbf{Verified Purchase}: A boolean indicator showing whether the product was purchased through Amazon, confirming the authenticity of the review.
\end{itemize}

Such user review data, which includes both non-text tabular data and text-based comment data, can well represent the complexity of the generated synthetic data. 

To generate synthetic review data, we select 50 samples from a dataset of software product reviews to train LLMs through prompting techniques. Each model is tasked with producing 300 synthetic entries. Claude demonstrates the most efficient data generation process, smoothly producing approximately 25 entries at a time without the need for re-prompting. In contrast, ChatGPT requires frequent reintroduction of the real data to maintain consistency in generating software related product reviews. Llama necessitates detailed prompting to accurately specify numeric data ranges and text formats; lacking precise instructions, it tends to shift focus to unrelated topics or generate data in incorrect formats.

A uniqueness verification is conducted to ensure the originality of each synthetic entry. Out of the requested 300 entries, Claude manages to produce 300 unique items. ChatGPT produces 292 unique entries. Llama, however, only produces 115, with the remainder being duplicates. These duplicates are removed prior to further evaluation to maintain the integrity of our data analysis.


\subsection{Models to evaluate}
In the experiment, we evaluate three advanced large language models (LLMs) to understand their performances in generating synthetic data. The models selected are Claude 3 Opus, ChatGPT 3.5, and Llama 2 13B, representing a mix of proprietary and open-source technologies, each with unique operational and accessibility characteristics.

Claude 3 Opus is a proprietary model developed by Anthropic. It is notable for its advanced natural language understanding and generation capabilities. This model operates on a subscription basis, requiring a monthly payment for access. The decision to include Claude in this work stems from its representation of cutting-edge, commercially available closed-source LLMs. Evaluating Claude allows us to assess the performance of pay-to-access models in synthetic data generation, providing insights into the value offered by such commercially structured tools.

ChatGPT 3.5, developed by OpenAI, is another proprietary model but is offered free of charge. It has gained significant attention for its robust performance in a variety of natural language processing tasks. The inclusion of ChatGPT 3.5 enables us to compare a no-cost proprietary model against its paid counterparts, offering a perspective on how accessibility influences the quality and utility of synthetic data generation, especially in academic and low-resource settings.

Llama 2 13B, an open-source model developed by Meta, stands out due to its accessibility and the transparency of its development process. As an open-source LLM, Llama allows researchers complete access to tweak its parameters and training methods. Including Llama in our evaluation provides a comprehensive view of how open-source models stack up against proprietary models in generating synthetic data, which is crucial for understanding the broader ecosystem of data generation technologies.

By examining these three models, we aim to uncover the nuances of synthetic data generation across different model architectures. This comparative analysis helps us delineate the performance differentials and potentially identify the optimal configurations and settings for each type of model in the domain of synthetic data generation.

\subsection{Evaluation Results}


\subsubsection{Fidelity}
Table 1 presents the fidelity evaluation of the non-text tabular data within our synthetic datasets. All three models achieve a 100\% score in structure preservation, demonstrating their ability to maintain the column names in the synthetic datasets relative to the real data. In terms of data integrity, Claude scores the highest. ChatGPT and Llama receive lower scores due to their propensity to generate duplicate titles. Claude and ChatGPT exhibit higher scores for column shape, attributed to their capabilities of preserving the distribution shape of most columns. Their scores are affected by the predominance of zero values in the 'helpful vote' category, which deviates from the real data. Llama scores the lowest, not only displaying zeros in all 'helpful vote' fields but also inaccurately representing most 'verified purchase' values as false, contrary to the real data, which predominantly features verified purchases.

\begin{table}[ht]
\centering
\begin{tabular}{@{}lccc@{}}
\toprule
\textbf{Metric}                      & \textbf{Claude} & \textbf{ChatGPT} & \textbf{Llama} \\
\midrule
Structure Preserving Score           & 100\%           & 100\%            & 100\%           \\
Data Integrity Score                  & 98.4\%          & 93.9\%           & 87.59\%         \\
Column Shapes Score                  & 80.92\%         & 80.97\%          & 62.29\%         \\
\bottomrule
\end{tabular}
\vspace{10pt}
\caption{Fidelity Evaluation Results on Non-Text Tabular Data}
\label{tab:fidelity_evaluation}
\end{table}

Table 2 presents the text analysis results comparing three synthetic data sets against real data. All models successfully capture the predominant sentiment found within the review data. Claude not only aligns closely with the top three keywords and sentiment-related words of the real data but also closely mirrors the original reviews' writing style. Although the other two models produce somewhat similar keywords, the significant distinction lies in the average length of the reviews. Claude maintains an average review length most akin to that of the real data, while the other models generate significantly shorter reviews on average. Notably, ChatGPT initially produces lengthy reviews but the length decreases over time. Llama, on the other hand, struggles to generate extended content without losing coherence. Overall, Claude excels in preserving the underlying thematic and stylistic nuances characteristic of the real dataset.

\begin{table}[ht]
\centering
\label{tab:text_analysis}
\begin{tabular}{@{} >{\raggedright\arraybackslash}p{3.5cm} >{\centering\arraybackslash}p{2.8cm} >{\centering\arraybackslash}p{2.8cm} >{\centering\arraybackslash}p{2.8cm} >{\centering\arraybackslash}p{2.8cm} @{}}
\toprule
\textbf{Metric} & \textbf{Claude} & \textbf{ChatGPT} & \textbf{Llama} & \textbf{Real} \\ \midrule
Sentiment Distribution & Positive (82\%) & Positive (78.42\%) & Positive (75.86\%) & Positive (82\%) \\
Top 3 Keywords & app, features, like & game, highly, app & product, highly, app & app, game, like \\
Top 3 Sentiment Words & great, more, easy & reliable, addictive, flashlight & great, more, much & good, sure, easy \\
Average length (words) & 40.48 & 16.55 & 18.69 & 59.26 \\
\bottomrule
\end{tabular}
\vspace{10pt}
\caption{Text Analysis Results}
\end{table}

\subsubsection{Utility}

Table 3 displays the utility evaluation results for synthetic data. While the sentiment classification model trained on real data achieves the highest accuracy, the models trained on synthetic data also exhibit comparable accuracy and similar Mean Absolute Error (MAE) values. This suggests that, despite varying levels of fidelity, all models effectively capture the relationship between review text data and user-provided ratings. Given that the training involved no more than 300 data samples, we anticipate that accuracy could further improve with the inclusion of more training data. Additionally, with an expanded dataset, the utility performance of each synthetic data might show greater divergence.

\begin{table}[ht]
\centering
\label{tab:evaluation_results}
\begin{tabular}{@{}lcc@{}}
\toprule
\textbf{Data Type} & \textbf{MAE} & \textbf{Accuracy} \\ \midrule
Claude Synthetic    & 1.2929       & 67.68\%                 \\
ChatGPT Synthetic   & 1.2041       & 67.35\%                 \\
Llama Synthetic    & 1.4151       & 62.26\%                 \\
Real Data           & 1.3019       & 67.92\%                 \\
\bottomrule
\end{tabular}
\vspace{10pt}
\caption{Utility Evaluation Results}
\end{table}

\subsubsection{Privacy}

Table 4 presents the privacy evaluation results, showcasing the accuracy of the trained Membership Inference Attack (MIA) models \cite{carlini2022membership}. All three models demonstrate very high successful rates, which in the context of MIA, suggests a higher probability of information leakage. This high successful rate is likely due to the fact that categorical features such as product IDs, parent product IDs, and user IDs are largely duplicated within each synthetic dataset. This duplication enables the models to re-identify data successfully.

These results imply that without additional fine-tuning, LLMs struggle to maintain the security and uniqueness of data in complex synthetic data generation tasks. Nonetheless, it is also conceivable that with more comprehensive datasets for training, these models could enhance their data generation capabilities, reducing the risk of privacy breaches.

\begin{table}[ht]
\centering
\label{tab:mia_model_accuracy}
\begin{tabular}{@{}lc@{}}
\toprule
\textbf{MIA Model} & \textbf{Successful Rate} \\ \midrule
Trained by Claude Synthetic & 91\% \\
Trained by ChatGPT Synthetic & 90\% \\
Trained by Llama Synthetic & 83\% \\
\bottomrule
\end{tabular}
\vspace{10pt}
\caption{Privacy Evaluation Results}
\end{table}

\section{Conclusion and Future Work}
In this paper, we propose a comprehensive evaluation framework for quantitatively assessing the fidelity, utility, and privacy of synthetic tabular data generated by LLMs. We demonstrated the effectiveness of the proposed framework by applying it to synthetic product review data generated by three state-of-the-art LLMs: ChatGPT, Claude, and Llama. This framework can contribute to the growing field of synthetic data evaluation. As LLMs continue to advance and generate increasingly realistic synthetic data, the proposed framework can serve as a foundation for future research and help ensure the responsible and effective use of synthetic data in real-world applications.

The proposed work also opens up opportunities for future research. First, while this study focuses on product review data, the evaluation framework can be extended to other domains and data types to assess. Second, the privacy evaluation component can be enhanced by incorporating additional privacy attacks and metrics to provide a more comprehensive assessment of privacy risks. Third, the trade-off between privacy and utility can be further explored by developing advanced privacy-preserving techniques that maintain high data utility while ensuring strong privacy guarantees.



\bibliographystyle{unsrt}  
\bibliography{references}

\begin{thebibliography}{10}

\bibitem{noteSynAI}
{Synthesis AI}.
\newblock {Synthetic Data Guide: Definition, Advantages, \& Use Cases}.
\newblock \url{https://synthesis.ai/synthetic-data-guide/}, 2023.
\newblock Accessed: 2024-04-20.

\bibitem{porter2023}
Alexis Porter.
\newblock Lessons learned from gdpr fines in 2023.
\newblock CPO Magazine, 2023.
\newblock [Online]. Available:
  \url{https://www.cpomagazine.com/data-protection/lessons-learned-from-gdpr-fines-in-2023/}
  [Accessed: 2024-04-20].

\bibitem{notePriManager}
{Data Privacy Manager}.
\newblock Meta hit with record €1.2b gdpr fine -- data privacy manager.
\newblock
  \url{https://dataprivacymanager.net/meta-hit-with-record-e1-2b-gdpr-fine/},
  2023.
\newblock Accessed: 2024-04-20.

\bibitem{Savage2023}
Neil Savage.
\newblock Synthetic data could be better than real data.
\newblock Nature, 2023.
\newblock Published in April.

\bibitem{Dilmegani2023}
Cem Dilmegani.
\newblock Synthetic data vs real data: Benefits, challenges in 2023.
\newblock \url{https://research.aimultiple.com/synthetic-data-vs-real-data/},
  2023.
\newblock Accessed: 2024-04-20.

\bibitem{VentureBeat2021}
VentureBeat.
\newblock 89\% of tech execs see synthetic data as a key to staying ahead.
\newblock
  \url{https://venturebeat.com/ai/89-of-tech-execs-see-synthetic-data-as-a-key-to-staying-ahead},
  2021.
\newblock Accessed: 2024-04-20.

\bibitem{Metinko2022}
Chris Metinko.
\newblock Synthetic data startups pick up more real cash.
\newblock
  \url{https://news.crunchbase.com/ai-robotics/synthetic-data-vc-funding-datagen-gretel-nvidia-amazon/},
  2022.
\newblock Accessed: 2024-04-20.

\bibitem{Yoon2019}
Jinsung Yoon, James Jordon, and Mihaela~v. Schaar.
\newblock Pate-gan: Generating synthetic data with differential privacy
  guarantees.
\newblock OpenReview, 2019.

\bibitem{Yoon2022}
Jinsung Yoon and Sercan~O. Arik.
\newblock Ehr-safe: Generating high-fidelity and privacy-preserving synthetic
  electronic health records.
\newblock
  \url{https://blog.research.google/2022/12/ehr-safe-generating-high-fidelity-and.html},
  2022.
\newblock Accessed: 2024-04-20.

\bibitem{Golshan2023}
Ali Golshan.
\newblock Gretel and google cloud partner on synthetic data.
\newblock \url{https://gretel.ai/blog/gretel-google-cloud-partnership}, 2023.
\newblock Accessed: 2024-04-20.

\bibitem{Barron2021}
Jenna Barron.
\newblock Microsoft introduces new tools for responsible ai.
\newblock
  \url{https://sdtimes.com/ai/microsoft-introduces-new-tools-for-responsible-ai/},
  2021.
\newblock Accessed: 2024-04-20.

\bibitem{Wiggers2021}
Kyle Wiggers.
\newblock Facebook quietly acquires synthetic data startup ai.reverie.
\newblock
  \url{https://venturebeat.com/business/facebook-quietly-acquires-synthetic-data-startup-ai-reverie/},
  2021.
\newblock Accessed: 2024-04-20.

\bibitem{Nvidia2023}
{Nvidia}.
\newblock Synthetic data for ai \& 3d simulation workflows.
\newblock \url{https://www.nvidia.com/en-us/omniverse/synthetic-data/}, 2023.
\newblock Accessed: 2024-04-20.

\bibitem{goodfellow2014generative}
Ian Goodfellow, Jean Pouget-Abadie, Mehdi Mirza, Bing Xu, David Warde-Farley,
  Sherjil Ozair, Aaron Courville, and Yoshua Bengio.
\newblock Generative adversarial nets.
\newblock {\em Advances in neural information processing systems}, 27, 2014.

\bibitem{vaswani2017attention}
Ashish Vaswani, Noam Shazeer, Niki Parmar, Jakob Uszkoreit, Llion Jones,
  Aidan~N Gomez, {\L}ukasz Kaiser, and Illia Polosukhin.
\newblock Attention is all you need.
\newblock {\em Advances in neural information processing systems}, 30, 2017.

\bibitem{soltana2017synthetic}
Ghanem Soltana, Mehrdad Sabetzadeh, and Lionel~C Briand.
\newblock Synthetic data generation for statistical testing.
\newblock In {\em 2017 32nd IEEE/ACM International Conference on Automated
  Software Engineering (ASE)}, pages 872--882. IEEE, 2017.

\bibitem{jordon2018pate}
James Jordon, Jinsung Yoon, and Mihaela Van Der~Schaar.
\newblock Pate-gan: Generating synthetic data with differential privacy
  guarantees.
\newblock In {\em International conference on learning representations}, 2018.

\bibitem{wan2017variational}
Zhiqiang Wan, Yazhou Zhang, and Haibo He.
\newblock Variational autoencoder based synthetic data generation for
  imbalanced learning.
\newblock In {\em 2017 IEEE symposium series on computational intelligence
  (SSCI)}, pages 1--7. IEEE, 2017.

\bibitem{li2023synthetic}
Zhuoyan Li, Hangxiao Zhu, Zhuoran Lu, and Ming Yin.
\newblock Synthetic data generation with large language models for text
  classification: Potential and limitations.
\newblock {\em arXiv preprint arXiv:2310.07849}, 2023.

\bibitem{heckerman2008tutorial}
David Heckerman.
\newblock A tutorial on learning with bayesian networks.
\newblock {\em Innovations in Bayesian networks: Theory and applications},
  pages 33--82, 2008.

\bibitem{frees1998understanding}
Edward~W Frees and Emiliano~A Valdez.
\newblock Understanding relationships using copulas.
\newblock {\em North American actuarial journal}, 2(1):1--25, 1998.

\bibitem{bao2017cvae}
Jianmin Bao, Dong Chen, Fang Wen, Houqiang Li, and Gang Hua.
\newblock Cvae-gan: fine-grained image generation through asymmetric training.
\newblock In {\em Proceedings of the IEEE international conference on computer
  vision}, pages 2745--2754, 2017.

\bibitem{zhang2017adversarial}
Yizhe Zhang, Zhe Gan, Kai Fan, Zhi Chen, Ricardo Henao, Dinghan Shen, and
  Lawrence Carin.
\newblock Adversarial feature matching for text generation.
\newblock In {\em International conference on machine learning}, pages
  4006--4015. PMLR, 2017.

\bibitem{xu2019modeling}
Lei Xu, Maria Skoularidou, Alfredo Cuesta-Infante, and Kalyan Veeramachaneni.
\newblock Modeling tabular data using conditional gan.
\newblock {\em Advances in neural information processing systems}, 32, 2019.

\bibitem{qi2024efficient}
Sibo Qi, Juan Chen, Peng Chen, Peian Wen, Xianhua Niu, and Lei Xu.
\newblock An efficient gan-based predictive framework for multivariate time
  series anomaly prediction in cloud data centers.
\newblock {\em The Journal of Supercomputing}, 80(1):1268--1293, 2024.

\bibitem{floridi2020gpt}
Luciano Floridi and Massimo Chiriatti.
\newblock Gpt-3: Its nature, scope, limits, and consequences.
\newblock {\em Minds and Machines}, 30:681--694, 2020.

\bibitem{wu2023comparative}
Sean Wu, Michael Koo, Lesley Blum, Andy Black, Liyo Kao, Fabien Scalzo, and Ira
  Kurtz.
\newblock A comparative study of open-source large language models, gpt-4 and
  claude 2: Multiple-choice test taking in nephrology.
\newblock {\em arXiv preprint arXiv:2308.04709}, 2023.

\bibitem{touvron2023llama}
Hugo Touvron, Louis Martin, Kevin Stone, Peter Albert, Amjad Almahairi, Yasmine
  Babaei, Nikolay Bashlykov, Soumya Batra, Prajjwal Bhargava, Shruti Bhosale,
  et~al.
\newblock Llama 2: Open foundation and fine-tuned chat models.
\newblock {\em arXiv preprint arXiv:2307.09288}, 2023.

\bibitem{tang2023does}
Ruixiang Tang, Xiaotian Han, Xiaoqian Jiang, and Xia Hu.
\newblock Does synthetic data generation of llms help clinical text mining?
\newblock {\em arXiv preprint arXiv:2303.04360}, 2023.

\bibitem{nguyen2015non}
Hoang-Vu Nguyen and Jilles Vreeken.
\newblock Non-parametric jensen-shannon divergence.
\newblock In {\em Machine Learning and Knowledge Discovery in Databases:
  European Conference, ECML PKDD 2015, Porto, Portugal, September 7-11, 2015,
  Proceedings, Part II 15}, pages 173--189. Springer, 2015.

\bibitem{dziugaite2015training}
Gintare~Karolina Dziugaite, Daniel~M Roy, and Zoubin Ghahramani.
\newblock Training generative neural networks via maximum mean discrepancy
  optimization.
\newblock {\em arXiv preprint arXiv:1505.03906}, 2015.

\bibitem{snoke2018general}
Joshua Snoke, Gillian~M Raab, Beata Nowok, Chris Dibben, and Aleksandra
  Slavkovic.
\newblock General and specific utility measures for synthetic data.
\newblock {\em Journal of the Royal Statistical Society Series A: Statistics in
  Society}, 181(3):663--688, 2018.

\bibitem{dwork2009complexity}
Cynthia Dwork, Moni Naor, Omer Reingold, Guy~N Rothblum, and Salil Vadhan.
\newblock On the complexity of differentially private data release: efficient
  algorithms and hardness results.
\newblock In {\em Proceedings of the forty-first annual ACM symposium on Theory
  of computing}, pages 381--390, 2009.

\bibitem{sweeney2002k}
Latanya Sweeney.
\newblock k-anonymity: A model for protecting privacy.
\newblock {\em International journal of uncertainty, fuzziness and
  knowledge-based systems}, 10(05):557--570, 2002.

\bibitem{machanavajjhala2007diversity}
Ashwin Machanavajjhala, Daniel Kifer, Johannes Gehrke, and Muthuramakrishnan
  Venkitasubramaniam.
\newblock l-diversity: Privacy beyond k-anonymity.
\newblock {\em Acm transactions on knowledge discovery from data (tkdd)},
  1(1):3--es, 2007.

\bibitem{shokri2017membership}
Reza Shokri, Marco Stronati, Congzheng Song, and Vitaly Shmatikov.
\newblock Membership inference attacks against machine learning models.
\newblock In {\em 2017 IEEE symposium on security and privacy (SP)}, pages
  3--18. IEEE, 2017.

\bibitem{patki2016synthetic}
Neha Patki, Roy Wedge, and Kalyan Veeramachaneni.
\newblock The synthetic data vault.
\newblock In {\em 2016 IEEE international conference on data science and
  advanced analytics (DSAA)}, pages 399--410. IEEE, 2016.

\bibitem{berger2014kolmogorov}
Vance~W Berger and YanYan Zhou.
\newblock Kolmogorov--smirnov test: Overview.
\newblock {\em Wiley statsref: Statistics reference online}, 2014.

\bibitem{carlini2022membership}
Nicholas Carlini, Steve Chien, Milad Nasr, Shuang Song, Andreas Terzis, and
  Florian Tramer.
\newblock Membership inference attacks from first principles.
\newblock In {\em 2022 IEEE Symposium on Security and Privacy (SP)}, pages
  1897--1914. IEEE, 2022.

\bibitem{hou2024bridging}
Yupeng Hou, Jiacheng Li, Zhankui He, An~Yan, Xiusi Chen, and Julian McAuley.
\newblock Bridging language and items for retrieval and recommendation.
\newblock {\em arXiv preprint arXiv:2403.03952}, 2024.

\end{thebibliography}

\end{document}